`

# Molecular CT: Unifying Geometry and Representation Learning for Molecules at Different Scales

Jun Zhang[1], Yao-Kun Lei[2,3], Yaqiang Zhou[4], Yi Isaac Yang[2,*] and Yi Qin Gao[1,2,3,5,6,*]

[1] *Changping Laboratory, Beijing 102200, China*

[2] *Institute of Systems and Physical Biology, Shenzhen Bay Laboratory, 518055 Shenzhen, China*

[3] *Beijing National Laboratory for Molecular Sciences, College of Chemistry and Molecular Engineering, Peking University, 100871 Beijing, China.*

[4] *Huawei Technologies Co., Ltd., 518192 Shenzhen, China*

[5] *Beijing Advanced Innovation Center for Genomics, Peking University, 100871 Beijing, China.*

[6] *Biomedical Pioneering Innovation Center, Peking University, 100871 Beijing, China.*

[*] Correspondence should be sent to yangyi@szbl.ac.cn (Y.I.Y) or gaoyq@pku.edu.cn (Y.Q.G).

## Abstract

Deep learning is changing many areas in molecular physics, and it has shown great potential to deliver new solutions to challenging molecular modeling problems. Along with this trend arises the increasing demand of expressive and versatile neural network architectures which are compatible with molecular systems. A new deep neural network architecture, Molecular Configuration Transformer (Molecular CT), is introduced for this purpose. Molecular CT is composed of a relation-aware encoder module and a computationally universal geometry learning unit, thus able to account for the relational constraints between particles meanwhile scalable to different particle numbers and invariant with respect to the trans-rotational transforms. The computational efficiency and universality make Molecular CT versatile for a variety of molecular learning scenarios and especially appealing for transferable representation learning across different molecular systems. As examples, we show that Molecular CT enables representational learning for molecular systems at different scales, and achieves comparable or improved results on common benchmarks using a more light-weighted structure compared to baseline models.

## I. INTRODUCTION

In recent years, artificial intelligence starts to transform many areas in science. Particularly, connectionist models or artificial neural networks (ANNs), are undergoing a resurgence in the form of *deep learning* by stacking (usually much) more than one hidden layer in ANNs and giving rise to models in relatively "deep" architectures.[1, 2] Compared to other functional approximators like polynomials, ANNs exhibit a better trade-off between expressivity and tractability, and they are more scalable when dealing with high-dimensional data, hence, can be very useful in a variety of tasks in molecular physics. For example, adopting proper ANNs operating on the electrons and atom nuclei, deep learning has delivered new solutions to the Schrodinger equation,[3] which is a long-standing challenge in quantum physics. Similarly, given the exquisitely and special-purposely designed ANNs, deep learning can help address other molecular modeling tasks[4] beyond the quantum chemistry, such as predicting protein structures,[5, 6] coarse graining bio-macromolecules,[7] kinetic modeling[8, 9] and enhanced sampling of many-particle Hamiltonian systems,[10, 11] *etc*.

However, since the inductive biases of atomistic and molecular systems differ from those of images or texts, the model architecture suitable for molecular systems is very different from traditional neural network models adopted in computer vision (CV) or natural language processing (NLP). Particularly, the properties (e.g., the energy or forces) of the atoms and molecules can be regarded as functions of the identities and the spatial positions of the particles, and these functions are known to preserve the reference invariance (or trans-rotational equivariance) as well as other symmetries of the many-particle systems.[4] Therefore, a deep neural network can be adopted to describe atomistic or molecular systems only if it is able to preserve these inductive biases.

During the past decade, the development of such ANN models has been spearheaded by researchers aiming to fit the *ab initio* potential energy surface (PES) of atomistic systems.[4, 12] The PES is a function mapping the types and positions of a collection of atoms into a scalar energy, which in principle can be solved from the Schrodinger equation. However, solving Schrodinger equation can be extremely time-consuming in real cases, so a surrogate function approximating the solution of Schrodinger equation which can be evaluated instantaneously is highly desired. The parametric ANNs are thus appealing for this specific purpose. The pioneering work along this line can be dated back to the Behler-Parrinello neural network (BPNN).[13] In BPNN, the chemical environment of each atom is transformed by specific symmetric (and trans-rotationally invariant) functions or descriptors called atomic environment vectors (AEVs) which are later used as input to a multilayer feed-forward network (FFN). BPNN has been deployed to approximate *ab initio* PES of various material systems, and its success catalyzed the birth of an expanded family of descriptor-based models, including ANI[14] and TensorMol[15] as well as DPMD[16], *etc.* Like all the other descriptor-based machine learning models, the performance of these models strongly depends on the selection of descriptors, or the *feature engineering* process. The expressivity of these atomistic learning models is limited once the feature sets are fixed, and transferable learning across different systems of varying sizes and/or components becomes a challenge. For instance, in BPNN, when the heterogeneity (i.e., types of atoms) of the system grows, the number of necessary input descriptors rapidly increases. Nevertheless, for a specific system given properly selected descriptors, both neural network models of this kind and non-parametric models (e.g., Gaussian processes) could perform well.[12,17]

On the other hand, with the advent of massage-passing neural network (MPNN)[18], machine learners began to approach the atomistic learning challenge from a relatively orthogonal view. Different from the descriptor-based models, in MPNN, the positions of the atoms are directly used as input to the neural network. Extracting a meaningful representation for the chemical environment of atoms is treated as a machine learning problem on its own. In other words, MPNN performs representation learning prior to predicting the properties. Such a representation learning is mostly based on graph neural networks (GNNs), because GNNs can naturally capture the useful priors for the molecular systems such as the permutation invariance.[4, 19] Models of this category include DTNN[20], SchNet[21], PhysNet[22] and DimeNet[23], to name only a few. From this respect, advancing from descriptor-based to GNN-based models largely resembles the paradigm-shift in CV research after the convolutional neural network (CNN) was invented. Compared to descriptor-based models, the GNN-based ones are able to adapt the description of heterogeneous chemical environments according to the training data, thus can lead to a better performance and transferability in certain contexts.[22,23]

Despite the increasing number of variants, the usage of these existing neural network models is mostly limited to atomistic learning, where the atoms are treated as “point cloud” in a quantum mechanical (QM) picture, that is, the models are invariant to the exchange of positions between two identical atoms. While in a more common setting, for many molecular systems, we are also concerned with some relational constraints between particles in addition to their spatial geometry, for instance, the amino acid sequence of proteins, the base sequence of nucleic acids, and the inter-connected springs of ring polymers, etc. Such representation of molecular systems defined by both the particle positions and their relational constraints is typically known as molecular mechanics (MM).[24] The relational constrains break the permutation invariance of GNN-based atomistic learning models, and it is non-trivial to derive transferable descriptors to encode such constraints for the descriptor-based models. Consequently, the development of proper models enabling learning at both QM and MM level still lags behind.

In this paper, we unified the treatment of atoms and molecules with/without relational constrains based on *dual representations* of molecular systems: one regarding the geometry (i.e., the spatial positions of the constituting particles like atoms), the other regarding the relational constraints between the particles (like the connectivity between atoms or the sequence of amino-acid residues). On the basis of such dual-representation framework:

1) We proposed a unified representation for atomistic and/or molecular systems using chemistry-based graph featurization, so that we can model a variety of many-particle systems and learn objectives on various scales using the same neural network architecture.

2) We developed an expressive many-body operator for geometric learning inspired by Transformer[25], a state-of-the-art deep learning architecture developed in machine learning community, which allows efficient representation learning of atomistic and/or molecular systems.

3) We assembled a universal (Turing complete) computational module, Neural Interaction Unit, which enables learning of transferable features for arbitrary atomistic and/or molecular systems using a single set of trainable parameters.

## II. METHODS & RELATED WORK

### 1. Unified graph featurization of many-particle systems with dual representations

Without loss of generality, we utilized *dual representations* to represent systems consisting of many particles: The coordinates of the constituting particles, and the relational constraints between the particles (Fig. 1). Since most molecular systems can be fully specified by these dual representations, we propose to treat them as structured graphs consisting of nodes (or vertices) and edges. In this featurized graph, the essential information of the particles (e.g., the types of which) is encoded into the nodes; While the edges connecting the nodes represent the relative positions as well as the possible relational constrains between the two particles. In the special case where no relational edges are present, the dual representations of atomistic systems reduce to the one commonly used in QM atomistic learning. Depending on the type of the relational edges, this graph can be either directed or undirected. This data structure enables us to represent atoms, molecules, proteins and even coarse-grained many-particle models in a unified manner.

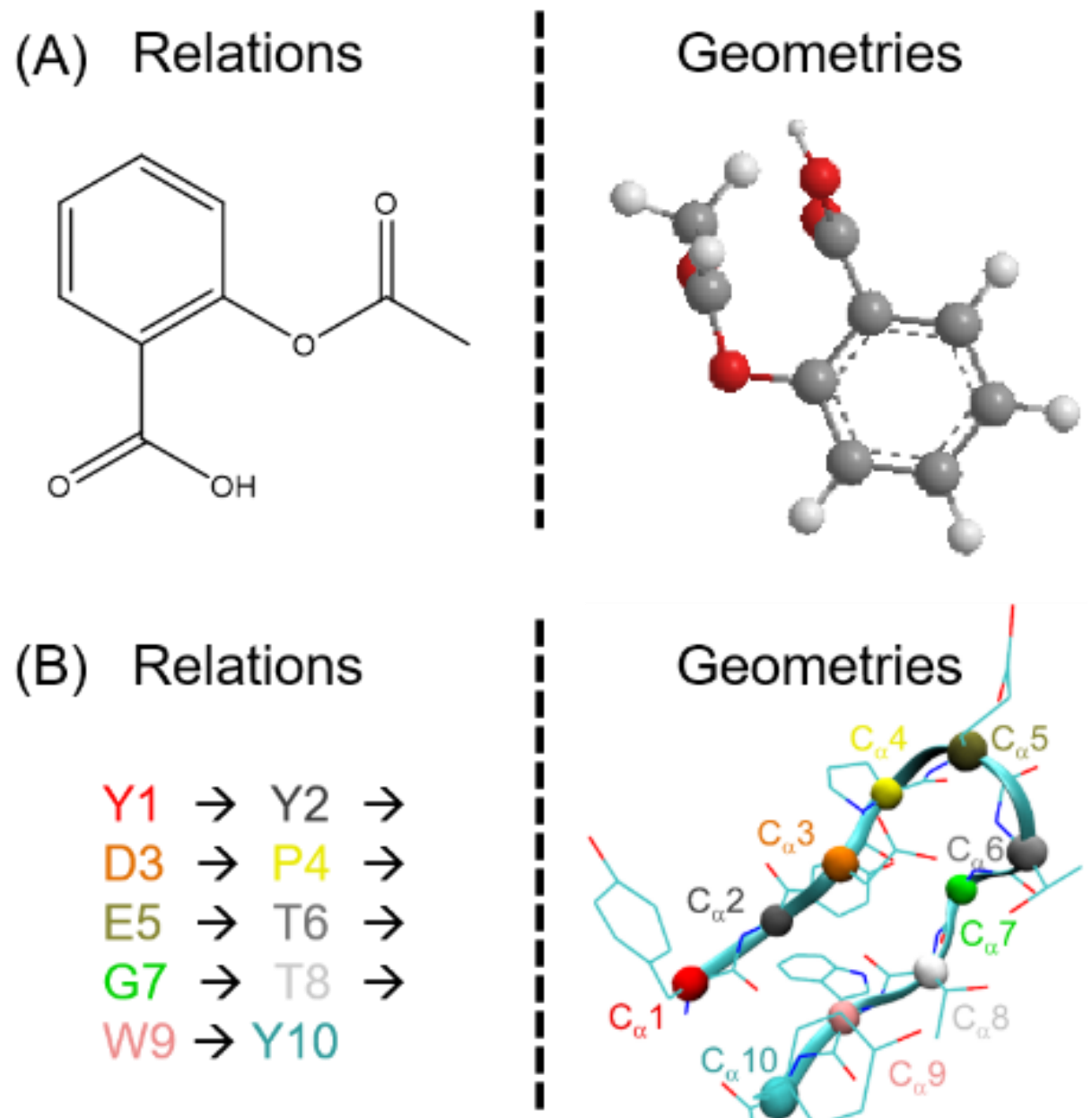

**FIGURE 1**. Dual representations of molecules. (A) Aspirin, represented by the molecular graph (left) and 3D configuration (right). Carbon, hydrogen and oxygen are colored in gray, white and red, respectively. (B) Chignolin protein, represented by the sequence of amino acids (left) and positions of carbon-$\alpha$ atoms (right) which are shown in beads colored according to the sequence on the left panel; The omitted side chains are shown by thin lines in the background.

The initial feature vector for the $i$-th node, denoted by $\mathbf{z}_i$, should encode the information disambiguating the identity of the $i$-th particle, in analogy to the word embedding[26] procedure in NLP. Besides, $\mathbf{z}_i$ should be a general and transferable representation which can be shared by a specific domain of many-particle systems. For instance, in QM-level learning, $\mathbf{z}_i$ can be a 1 by $D$ row vector mapped from the atomic nuclear charge, where $D$ is the dimensionality of the latent embedding. While in even coarser-grained scales, a single coarse-grained particle may directly represent a residue of a protein (Fig. 1B), $\mathbf{z}_i$ then should encode the identity of the corresponding amino acid.

In line with the dual representations, the featurized edge between the $i$-th and $j$-th nodes contains two independent parts: One relational edge vector $\mathbf{v}_{ij}$ and one positional edge vector $\mathbf{e}_{ij}$. The relational edge vector $\mathbf{v}_{ij}$ accounts for the relational constraints (if any) between the $i$-th and $j$-th particles. For example, for organic molecules, which are commonly described by the corresponding molecular graphs, $\mathbf{v}_{ij}$ should encode the bonding connectivity between atoms (Fig. 1A). For proteins or DNA/RNAs, $\mathbf{v}_{ij}$ should describe the relative positions (including directions) of residues in a given sequence of amino acids or nucleotides (Fig. 1B). This can be done straightforwardly by mapping different types of relational constrains into (optimizable) edge vectors $\mathbf{v}_{ij} \in \mathbb{R}^{1\times d}$.[18]

On the other hand, the positional edge vector $\mathbf{e}_{ij}$ is constructed to encode the geometric information of molecules. To guarantee the overall model be invariant w.r.t. trans-rotations, we use the inter-particle distance $r_{ij} = |\mathbf{R}_i - \mathbf{R}_j|$ (where $\mathbf{R}_i$ denotes the Cartesian coordinates of the $i$-th particle) to represent the relative positional information between any two particles. Noteworthy, we choose $\log r_{ij}$ instead of $r_{ij}$ for edge featurization. This choice is motivated by the facts that learning linear transforms (as done by matrix multiplications in neural networks) on $\log r_{ij}$ is equivalent to learning power functions of $r_{ij}$, and that in physics many interactions are empirically expressed as power functions of the two-body distance $r_{ij}$, for instance, $r_{ij}^{-1}$ in the Coulomb potential. The logarithm transform also implicitly preserves the asymptotic singularity dictated by the Pauli exclusion principle that atoms (or other related physical particles) should not overlap (i.e., $r_{ij} \neq 0$). One could directly concatenate $\log r_{ij}$ into $\mathbf{e}_{ij}$. However, inserting a single scalar into a high-dimensional vector $\mathbf{e}_{ij}$ is not information efficient, and poses challenges for the neural network to utilize this piece of information. Inspired by the positional embedding technique introduced in Transformer,[25] we expanded $\log r_{ij}$ by a radial basis functions (Eq. (1)) into a $d$-dimensional row vector:

$$
\begin{aligned}
\mathbf{e}(r_{ij}) &= \left[e_1(r_{ij}), e_2(r_{ij}), \ldots, e_d(r_{ij})\right] \\
e_k(r_{ij}) &= \exp\left[-\frac{\left(\log r_{ij} - \mu_k\right)^2}{2\sigma^2}\right]
\end{aligned}
\qquad (1)
$$

where $\mu_k$'s are evenly spaced in $[\log r_{\text{min}}, \log r_{\text{cut}}]$, and $\sigma$ is a hyper-parameter determining the spatial resolution of the model. As Fig. 2 shows, given the lower bound $r_{\text{min}}$ and the cutoff distance $r_{\text{cut}}$, the radial basis expansions of $\log r_{ij}$ not only encode the distance information into $\mathbf{e}_{ij}$ in a continuous and information-rich manner, but also "blurs" the relative positions of particles which are farther apart than the cutoff. This corresponds to a common physics intuition that the state of a particle is largely determined by its local environment. Note that similar radial basis expansion techniques were also adopted by other models like SchNet and PhysNet, but in these models, it is the distance $r_{ij}$ that is directly expanded by certain basis functions.

For a system consisting of $N$ particles, we first transform their initial node vectors $\{\mathbf{z}_i\}_{i=1,\ldots,N}$ into relation-aware ones $\{\mathbf{n}_i\}_{i=1,\ldots,N}$ using the relational edge vectors $\{\mathbf{v}_{ij}\}_{i,j=1,\ldots,N}$ via the Relational Molecule Encoder (see Section II.2), then gradually refine the representations of the nodes using the positional edge vectors $\{\mathbf{e}_{ij}\}_{i,j=1,\ldots,N}$ via the Neural Interaction Unit (see Section II.4).

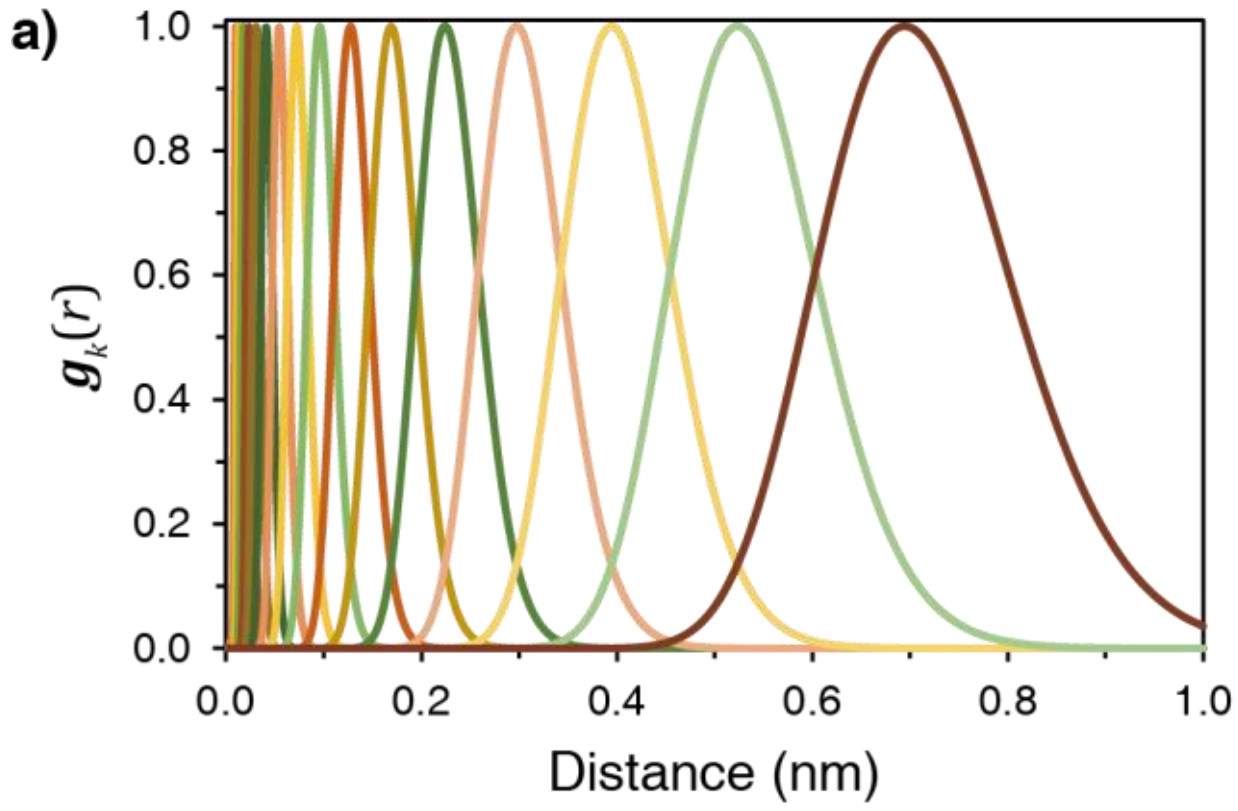

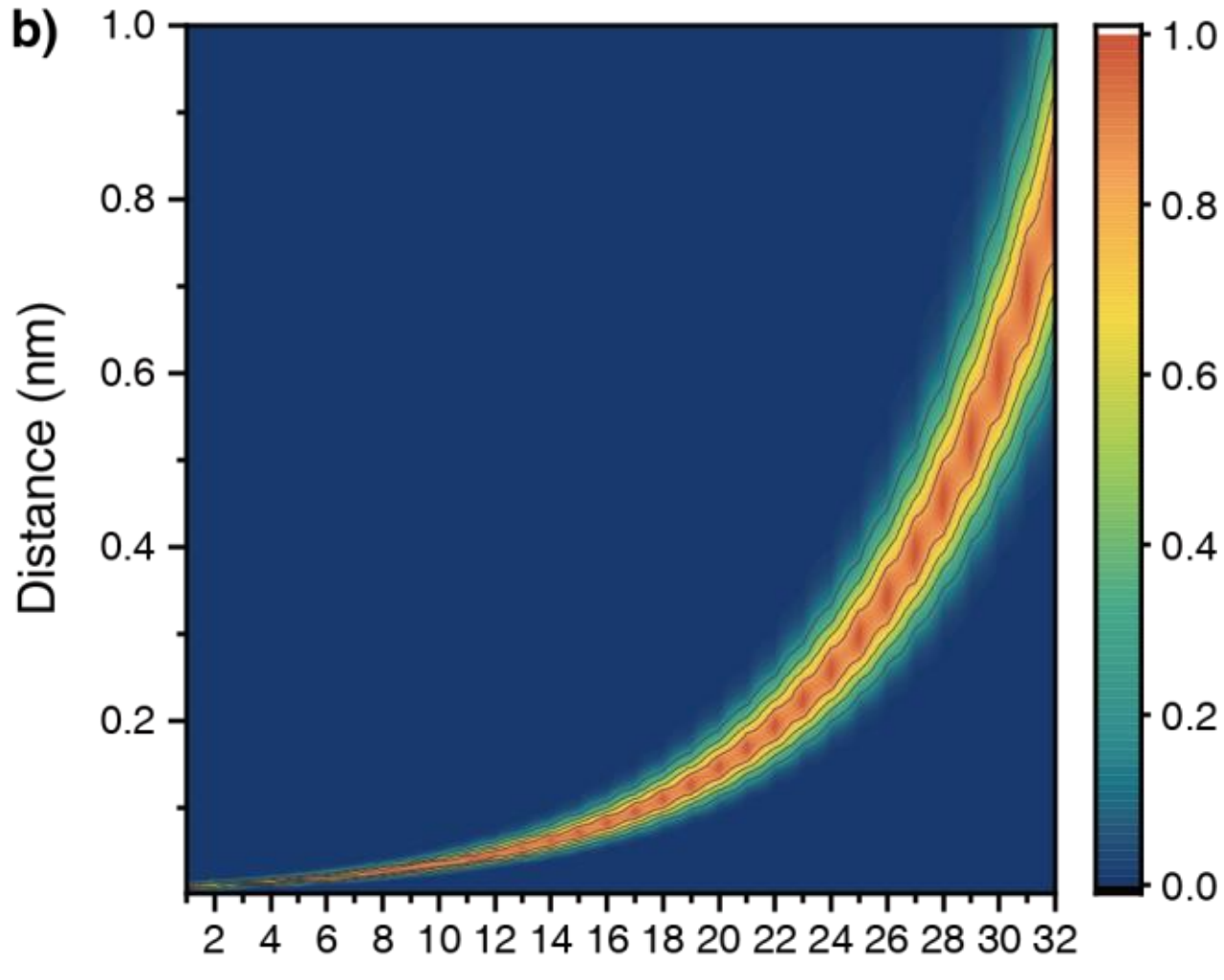

**FIGURE 2**. Positional edge featurization based on logarithm-distance. (A) RBFs of $\log r_{ij}$ at different $\mu_k$ shown in different colors. (B) The $d$-dimensional ($d$=32) positional edge vector $\mathbf{e}_{ij}$ is a smooth vector function of $r_{ij}$.

### 2. Relational Molecule Encoder (RME) for relation-aware representation learning

RME is designed to transform the relation-independent node vectors $\{\mathbf{z}_i\}_{i=1,\ldots,N}$ into relation-aware ones $\{\mathbf{n}_i\}_{i=1,\ldots,N}$ prior to the subsequent geometric learning, and we exploit a Transformer-like architecture for this purpose.

Attention mechanism[27] is a widely adopted technique in modern deep learning, and it has outperformed traditional models like CNNs and recurrent neural networks (RNNs) in a variety of machine learning tasks. The multi-head attention, a variant of attention mechanism, is the bedrock of the famous Transformer model.[25] Specifically, given a row query vector $\mathbf{Q} \in \mathbb{R}^{1\times D}$, $N$ row-batched key vectors $\mathbf{K} \in \mathbb{R}^{N\times D}$ and value vectors $\mathbf{V} \in \mathbb{R}^{N\times D}$ (note that the row-to-row correspondence between the key-value pairs is fixed), a single-head scaled dot-product attention (Fig. 3A) is a non-linear operation defined by Eq. (2),

$$\begin{aligned} \text{Attention}(\mathbf{Q},\mathbf{K},\mathbf{V}) &= \boldsymbol{\alpha}\mathbf{V} \\ \boldsymbol{\alpha} &= \text{softmax}\left(\frac{\mathbf{Q}\mathbf{K}^\top}{\sqrt{D}}\right) \end{aligned} \tag{2}$$

The softmax operation in Eq. (2) generates a set of normalized attention coefficients $\boldsymbol{\alpha} \in \mathbb{R}^{1\times N}$ for the query $\mathbf{Q}$ to sum over the values $\mathbf{V}$ given the keys $\mathbf{K}$. Note that this attention operation generates a new row vector in the same shape as $\mathbf{Q}$, and the attention operation is invariant w.r.t. the permutation of the rows in the batched key-value vector pairs. On the basis of Eq. (2), the multi-head attention (MHA; Fig. 3B) with $k$ heads is defined as,

$$\begin{aligned} \text{MHA}(\mathbf{Q},\mathbf{K},\mathbf{V}) &= [\text{head}_1,\ldots,\text{head}_k]\mathbf{W}^{(O)} \\ \text{head}_i &= \text{Attention}\left(\mathbf{Q}\mathbf{W}_i^{(Q)},\mathbf{K}\mathbf{W}_i^{(K)},\mathbf{V}\mathbf{W}_i^{(V)}\right) \end{aligned} \tag{3}$$

where $\mathbf{W}_i^{(Q)} \in \mathbb{R}^{D\times D/k}$, $\mathbf{W}_i^{(K)} \in \mathbb{R}^{D\times D/k}$, $\mathbf{W}_i^{(V)} \in \mathbb{R}^{D\times D/k}$ and $\mathbf{W}^{(V)} \in \mathbb{R}^{D\times D}$ are optimizable parameter matrices used for the affine projections.

For every initial node vector $\mathbf{z}_i \in \mathbb{R}^{1\times D}$, the query, keys and values are calculated according to Eq. (4), where $\mathbf{W}^{(K_2)}, \mathbf{W}^{(V_2)} \in \mathbb{R}^{d\times D}$ are optimizable affine matrices acting on the edges. We then update the node embedding through Eq. (5) in order to propagate the relational information.

$$\begin{aligned} \mathbf{Q}_i &= \mathbf{z}_i \\ \mathbf{k}_{j|i} &= \mathbf{z}_j + \mathbf{v}_{ij}\mathbf{W}^{(K_2)} \\ \mathbf{v}_{j|i} &= \mathbf{z}_j + \mathbf{v}_{ij}\mathbf{W}^{(V_2)} \\ \mathbf{K}_i &= \left[\mathbf{k}_{1|i}^\top,\mathbf{k}_{2|i}^\top,\ldots,\mathbf{k}_{N|i}^\top\right]^\top \\ \mathbf{V}_i &= \left[\mathbf{v}_{1|i}^\top,\mathbf{v}_{2|i}^\top,\ldots,\mathbf{v}_{N|i}^\top\right]^\top \end{aligned} \tag{4}$$

$$\mathbf{n}_i = \mathbf{z}_i + \text{MHA}(\mathbf{Q}_i,\mathbf{K}_i,\mathbf{V}_i)^\top \tag{5}$$

For each node, a neighbor mask (Fig. 3A) is also applied to zero out the contribution of its non-connected particles to the computed attention coefficients $\boldsymbol{\alpha}$ (Eq. (2)). As in Transformer, the RME attaches a position-wise FFN sublayer (Fig. 3C) (a position-wise network operates on a single node at a time, shared by all the nodes) post to the MHA sublayer. Residual connections are implemented at both sublayers followed by layer normalization.[28] The MHA plus the position-wise FFN constitutes one building block for RME, and one RME can be made up of a number of such blocks (Fig. 3C).

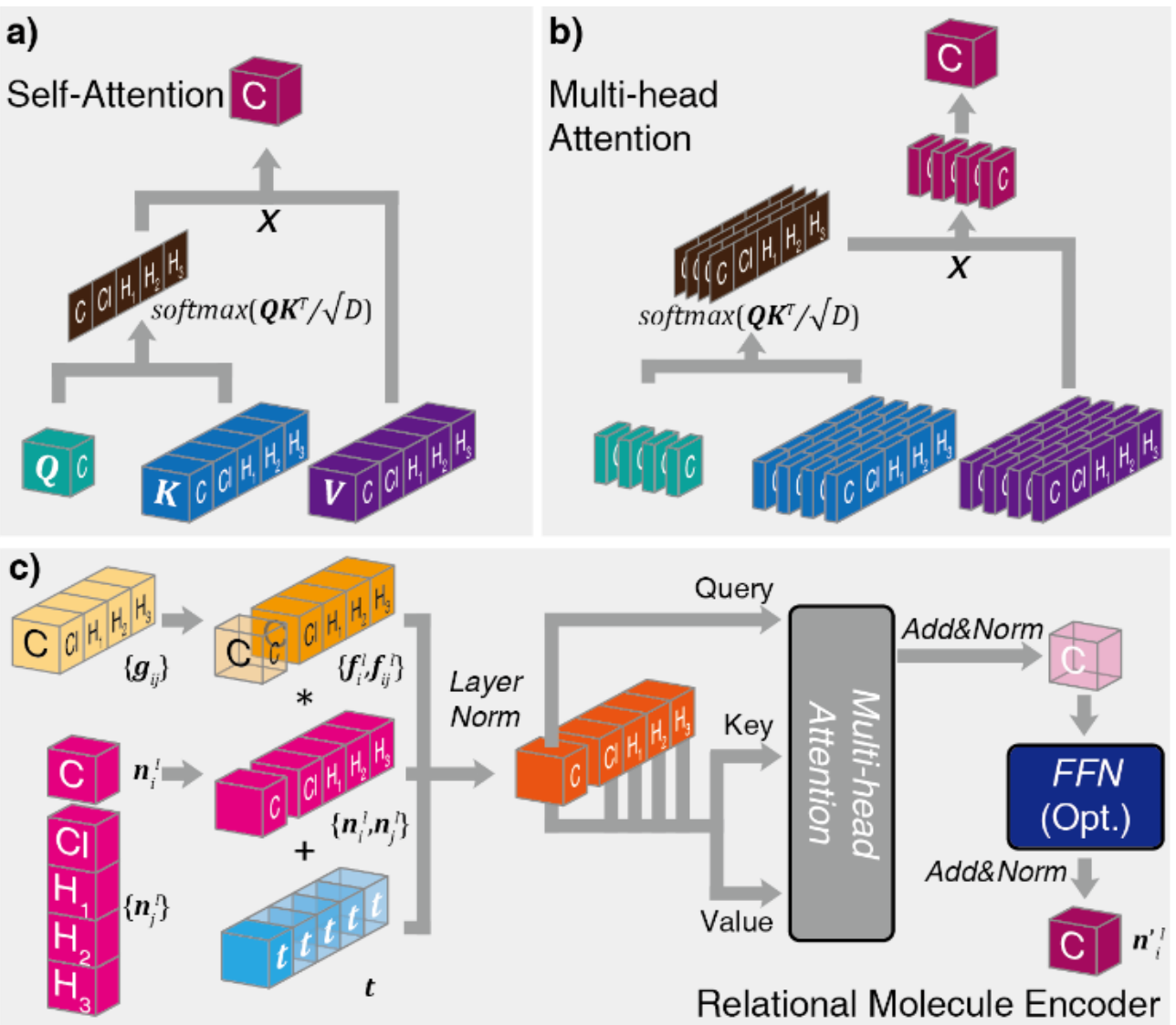

**FIGURE 3**. Illustrative computational graph for the Relational Molecule Encoder (RME). (A) Dot-product attention operation. (B) Multi-head attention operation containing $k$ heads. (C) Assembled RME containing $N$ building blocks.

Note that all the operations involved in RME, including MHA and layer-normalization, are independent of the particle number $N$ and the input order of particles, meaning that the same neural network can operate on systems containing different number of particles and preserve the permutation invariance. Besides, all the computation in RME is performed without the knowledge of geometric or structural information, therefore, RME alone can be applied to datasets containing only molecular graphs or sequences but no structural information. From this perspective, RME can serve as an alternative for other configuration-independent molecular learning models.

Given a set of initial node feature vectors, RME transforms them into relation-aware ones. While the initial node vectors can only distinguish particles by their identities or types (e.g., atomic nuclear charges), the output of RME can further disambiguate particles by the contexts and relations. For instance, the initial node vectors for an sp2 carbon atom and an sp3 carbon atom are identical, but after RME, the embedding vectors of the two carbon atoms can be different provided that these two carbons have different bond orders (or number of connections to other atoms). In many cases, such context-dependent and relation-aware embedding will be a good start for downstream learning task like predicting the molecular properties.

In next section, we will elaborate on how to incorporate structural information to further refine the node representations based on these relation-aware embedding vectors.

**3. Ego-Attention: An expressive many-body operator for geometric learning**

Notice that the relation-aware node embedding $\mathbf{n}_i$ does not contain structural information (if no additional many-body equivariant features are provided in the initial node vectors), and the positional edge vector $\mathbf{e}_{ij}$ contains merely two-body relative positions. However, many structure-dependent molecular properties are functions of the positions of many particles (like bond angles and torsions *etc.*). Therefore, the model has to learn to propagate structural information between nodes in order to approximate the many-body couplings between particles. In other words, for the sake of learning a useful representation for the particles, the model must be equipped with a *many-body operator* which propagates and integrates geometric information over many particles.

Indeed, many existing GNN-based atomistic learning models can be distinguished by their built-in many-body operators, and two important methods are worth mentioning. The first one is *message passing* (MP), formally summarized in MPNN paper and later extended by other variants.[18, 22, 23] In MPNN, "messages" (vectors) are generated for each node based on the latest node embedding using a shared neural network. These messages are then passed between every two neighboring nodes, depending on the positional edge vectors in between, and the node embedding gets updated by absorbing these messages. Note that the MP operation in each step is two-body in nature given that it is executed between every two neighboring particles. However, by iterating this MP procedure, each node is able to receive messages from nodes that are farther apart, and the information of many nodes gets convolved, effectively leading to a many-body operator. In practice, MP may not be expressive enough as a many-body operator unless sufficient number of iterations are invoked, which often entails heavy over-parametrization and may cause vanishing or exploding gradient issues. A recent variant of MPNN tried to integrate angular (three-body) features during message passing and observed improved performance.[23]

Another useful many-body operator is the *continuous-filter convolution* (CFC) originally proposed in SchNet.[21] In SchNet, a number of convolution filters (or kernels) are first generated by a neural network transform of pairwise distance features, based on which a convolution operation was performed over all nodes in analogy to the convolution operation in CNNs for pixel images. Compared to MP, CFC utilizes a relatively simple neural network architecture, and usually requires a smaller number of parameters. However, the core of CFC, i.e., the filter-generating network, is indeed a two-body operator, thus more than one CFC operation needs to be stacked, as in MP, in order to be truly "many-body". Besides, since CFC is defined as a generalized "convolution" over a spatial point cloud, it is not simply compatible with relational constraints, thus being limited to atomistic learning on the QM level. A recent study had to combine CFC with a separate GNN in order to deal with coarse-grained molecular systems.[7]

In the following, we propose a novel trans-rotation-invariant and permutation-invariant many-body operator on the basis of attention mechanism (Fig. 4A). In order to utilize the spatial information during attention, we perform *positional embedding* as in Transformer. In our model, the positional embedding of a node depends on the reference node, because in physics the position of a particle is meaningless without a reference. Specifically, once a node $\mathbf{n}_i$ is chosen as the reference, the positional embedding vectors $\{\mathbf{P}_{j|i}\}_{j=1,\ldots,N}$ for all the node $\{\mathbf{n}_j\}_{j=1,\ldots,N}$ with respect to $\mathbf{n}_i$ are computed through a linear transform (Eq. (6)),

$$\mathbf{P}_{j|i} = \mathbf{e}_{ij}\mathbf{W}^{(P)} + \mathbf{b}^{(P)} \tag{6}$$

where the positional edge vectors $\mathbf{e}_{ij}$ ($\mathbf{e}_{ii}$ is a constant vector for all $i$) are transformed; $\mathbf{W}^{(P)} \in \mathbb{R}^{d\times D}$ and $\mathbf{b}^{(P)} \in \mathbb{R}^{1\times D}$ are optimizable parameters. The position-embedded node vectors $\{\tilde{\mathbf{n}}_{j|i}\}_{j=1,\ldots,N}$ centric at the $i$-th node can be computed according to Eq. (7) where $\otimes$ denotes element-wise multiplication,

$$\tilde{\mathbf{n}}_{j|i} = \mathbf{n}_j \otimes \mathbf{P}_{j|i} \tag{7}$$

Similar relative positional embedding was also adopted in a Transformer variant which could deal with graph-structured data.[29] For the $i$-th node, we use the position-embedded vector $\tilde{\mathbf{n}}_{i|i} \in \mathbb{R}^{1\times D}$ as the query, while $\{\tilde{\mathbf{n}}_{j|i}\}_{j=1,\ldots,N}$ as both the keys and values for MHA (Eq. (8)). A pre-layer normalization[30] is implemented to keep the magnitude of the vectors proper for the dot-product attention, and the node vector $\mathbf{n}_i$ gets updated through Eq. (9),

$$\begin{aligned}\mathbf{Q}_i &= \mathrm{LayerNorm}\left(\tilde{\mathbf{n}}_{i|i}\right)\\ \mathbf{K}_i = \mathbf{V}_i &= \mathrm{LayerNorm}\left(\left[\tilde{\mathbf{n}}_{1|i}^{\top}, \tilde{\mathbf{n}}_{2|i}^{\top}, \ldots, \tilde{\mathbf{n}}_{N|i}^{\top}\right]^{\top}\right)\end{aligned} \tag{8}$$

$$\mathbf{n}_i = \mathbf{n}_i + \mathrm{MHA}\left(\mathbf{Q}_i, \mathbf{K}_i, \mathbf{V}_i\right)^{\top} \tag{9}$$

Intuitively, Eqs. (6-9) allow the $i$-th node to couple with other nodes through a self-centric view (i.e., with a self-centered relative positional embedding), therefore, we call this approach Ego-Attention (EA). Note that EA is performed for every node in the graph using the same set of learnable parameters, and a schematic computational graph for EA is shown in Fig. 4A.

Remarkably, thanks to the softmax in Eq. (2), EA is an intrinsic many-body operator. Therefore, unlike MP or CFC, even a single execution of EA could suffice to approximate complicated many-body interactions. EA also allows us to peek into the working mechanism of the model. For example, we may discover certain interaction patterns between particles by observing how a particle attends to other particles based on the learned attentional coefficients (Eq. (2)). Moreover, the computation of EA consists mainly of tensor multiplications which can be very efficient in modern parallel computing devices.

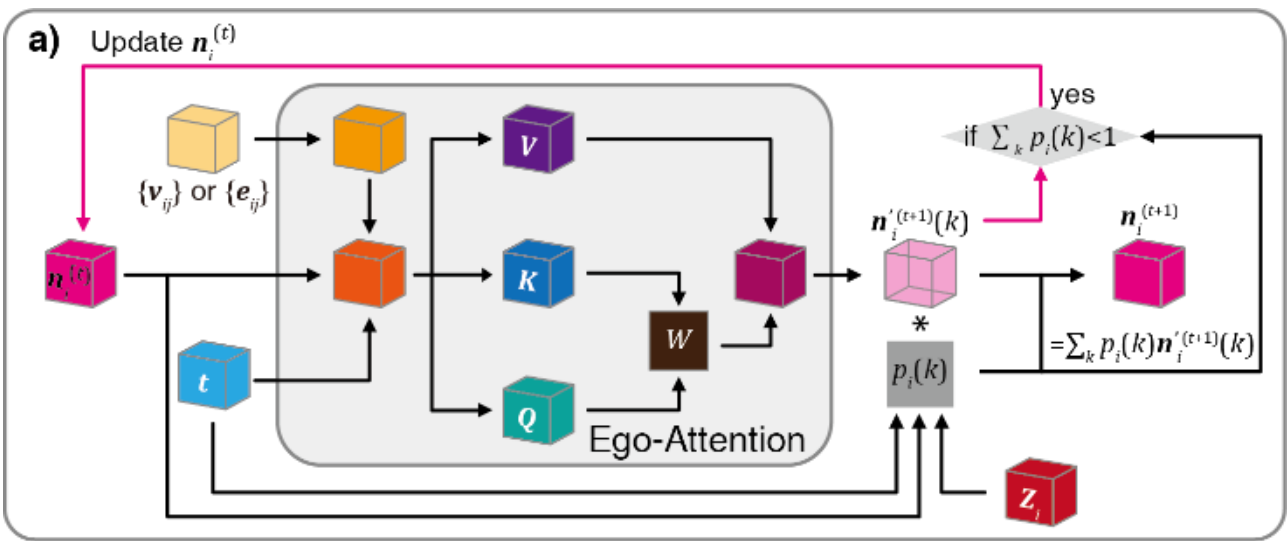

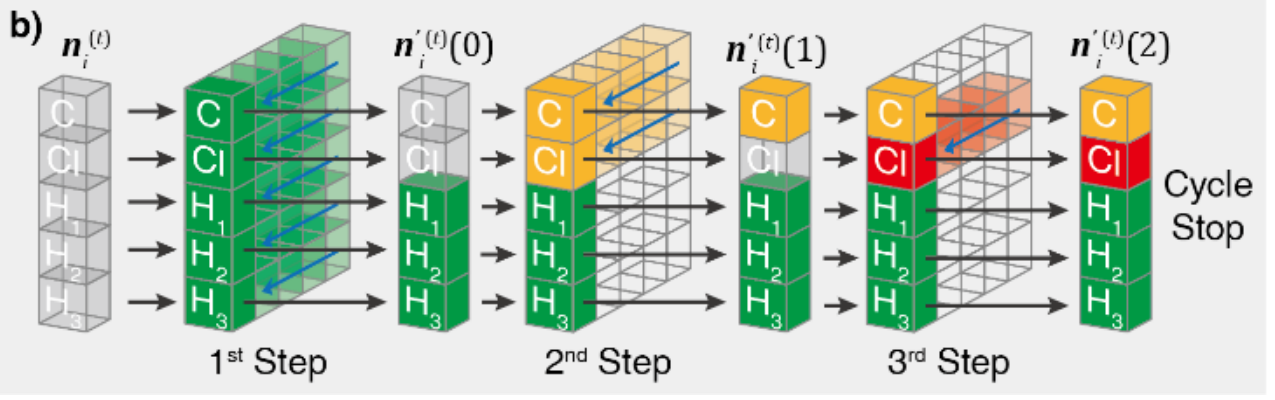

**FIGURE 4**. Illustrative computational graph for: (A) Ego Attention; (B) Computational Universal Neural Interaction Unit (NIU).

Even though EA allows one particle to attend to all the other particles in the system, in practice, it is usually useful to restrict the "horizon" of a particle by a cutoff. We can reduce the attentional coefficients between particles that are farther away than the cutoff distance to nearly zero in a continuous fashion. Specifically, we penalize the attention coefficient $\boldsymbol{\alpha}$ (Eq. (2)) from $\mathbf{n}_i$ to node $\mathbf{n}_j$ by a decay function with respect to their distance as adopted in BPNN.[13] Motivated by physics intuitions, the limited horizon assumption is widely adopted in atomistic learning, and it could help ameliorate overfitting meanwhile reducing the computational cost.

As in Transformer and RME, one may optionally include a position-wise FFN with a single hidden layer along with a residual connection to refine the node embedding post to EA:

$$\mathbf{n}_i = \mathbf{n}_i + \mathrm{FFN}\left(\mathrm{LayerNorm}\left(\mathbf{n}_i\right)\right) \tag{10}$$

Similar position-wise networks are required by other GNN-based atomistic learning models.[21-23] However, in our experiments, we found that Eq. (9) alone is so expressive that we can often forgo the use of Eq. (10), thus cutting the weight and computational cost of the overall model.

## 4. Computationally universal Neural Interaction Unit (NIU)

From another perspective, given a node embedding vector $\mathbf{n}_i^{(t)}$ and its associated edge vectors $\{\mathbf{e}_{ij}\}_{j=1,\ldots,N}$, equation (9) yields a rule of dynamics which evolves $\mathbf{n}_i^{(t)}$ to $\mathbf{n}_i^{(t+1)}$:

$$\mathbf{n}_i^{(0)} \xrightarrow{\text{EA}} \mathbf{n}_i^{(1)} \xrightarrow{\text{EA}} \dots \xrightarrow{\text{EA}} \mathbf{n}_i^{(T)} \tag{11}$$

In GNN-based models, usually a number of many-body operations (or interaction layers in some literature) need to be iterated (or stacked) for good performance. However, in many cases, the number of iterations or the stacked layers, $T$ in Eq. (11), through which $\mathbf{n}_i^{(0)}$ is evolved to $\mathbf{n}_i^{(T)}$, is a very sensitive hyper-parameter impacting the final performance. How to choose this hyper-parameter remains a common practical issue. Since the iterations effectively define a dynamics process for the node embedding vectors, in principle, $T$ should be large enough to allow every $\mathbf{n}_i^{(T)}$ reaches stationary (or the fixed) point.

However, the necessary computational complexity (or Kolmogorov complexity) of learning proper representations for particles may depend on the type of the particle and the specific context. Therefore, one may expect that the necessary $T$ should not be the same for different particles or systems. Consider the worst case that, if most particles in a system can reach stationary within a small $T_1$ steps, except that one particle requires $T_2 \gg T_1$ to get stationary, then all the other particles have to be computed up to a formidable $T_2$ iterations as well, which would cause severe computational waste in practice. More importantly, from the perspective of computer science, a model consisting of any finite and fixed $T$ interaction layers is not computationally universal (or not Turing complete).[31] This means that there always exists a system, the representation learning of which would lie beyond the capacity of such a model. This shortcoming may limit the ability for models to do transferable learning across different molecular systems.

To address these issues, we designed a computationally universal module, called Neural Interaction Unit (NIU) on the basis of EA. In a NIU, the iteration step $T$ in Eq. (11) is dynamically determined for each node during learning, and the parameters within a NIU are tied across all the iteration steps. This is achieved through Adaptive Computational Time,[32] a technique developed for RNNs but also adopted by the Universal Transformer.[31]

Specifically, in the NIU, we employ a position-wise pondering network which determines whether the computation should halt for the $i$-th node at $t$-th step based on $\mathbf{n}_i^{(t)}$. This pondering network is a single-layer FFN. The input to the pondering network is a time-embedded node vector, based on which the network predicts a halting probability $P_{\text{halt}}$ using a sigmoid output unit (Eq. (12)).

$$P_{\text{halt}}\left(\mathbf{n}_i^{(t)}\right) = \text{sigmoid}\left(\text{FFN}\left(\mathbf{n}_i^{(t)} \oplus \mathbf{T}^{(t)}\right)\right) \tag{12}$$

where $\oplus$ denotes element-wise addition, and the entries of the $D$-dimensional time-embedding vector $\mathbf{T}^{(t)}$ take the form in Eq. (13). Once the pondering network decides to halt the computation of the $i$-th node at $t$-th iteration, we will simply copy $\mathbf{n}_i^{(t)}$ to $\mathbf{n}_i^{(t')}$ for any afterward $t' > t$ till all the nodes have halted (Fig. 4C), hence allowing "early stopping" for those particles which are easier to learn.

$$\begin{aligned} \mathbf{T}_i^{(t)} &= \begin{cases} \sin\left(\omega_k t\right) & \text{if } i = 2k \\ \cos\left(\omega_k t\right) & \text{if } i = 2k+1 \end{cases} \\ \omega_k &= \frac{1}{10000^{2k/D}} \end{aligned} \tag{13}$$

Note that varying the number of computational iterations of a NIU after training is possible owing to the parameter tying and the time embedding technique. One can thus perform training with a moderate upper limit for the maximally allowed $T$ (which is equivalent to regularizing the Kolmogorov complexity of the model[32]), but remove this limitation during inference. Given sufficient memory, NIU is computationally universal, i.e., it belongs to the class of models that can be used to simulate any Turing machine as the famous Neural Turing Machine[33] and Differentiable Neural Computers[34]. One can assemble several different NIUs in stack or in parallel to achieve larger model capacity and expedite the training process.

In the final model, the RME and NIUs are assembled together to process both the relational constraints and structural information of a many-particle system (Fig. 5). We name this model architecture as the Molecular Configuration Transformer (Molecular CT) given its similarity to the Transformer model, and Molecular CT enables us to model a wide scope of many-particle systems in a unified manner.

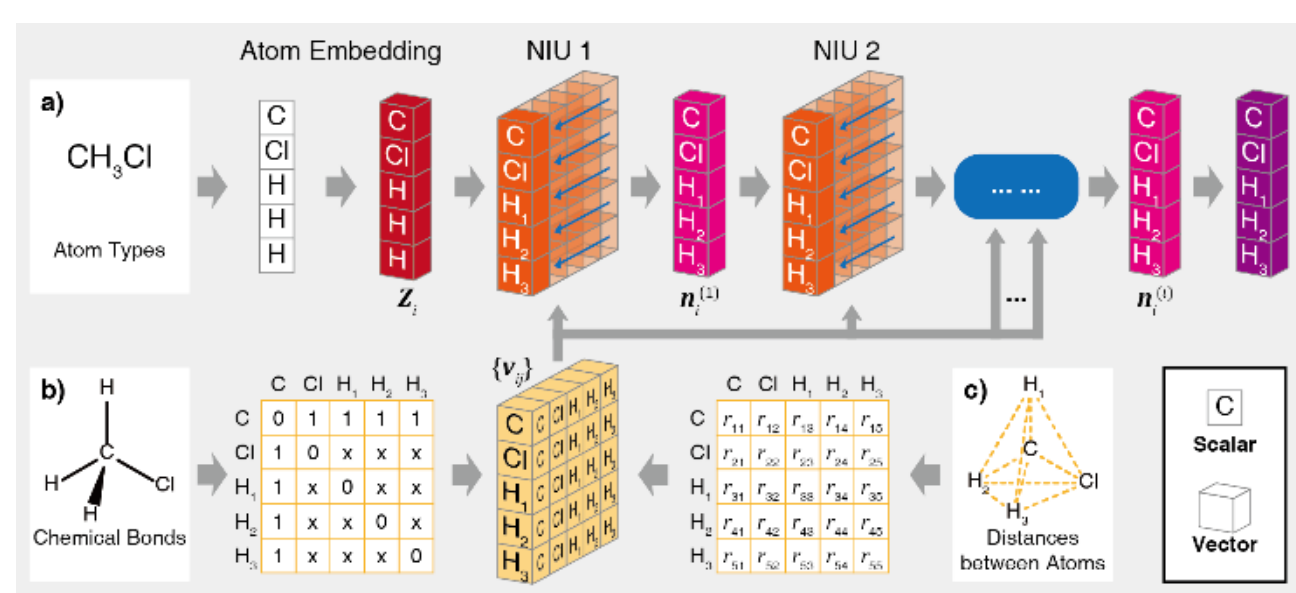

**FIGURE 5**. Illustrative computational graph of Molecular CT, which consists of the RME and $N$ stacked NIUs. Molecular CT operates on either the relational constraints or particle positions or both, and renders the embedding for each particle in a relation-aware and transrotation-invariant manner.

## 5. Representation learning on multiple levels

As a GNN-based model, Molecular CT is able to perform representation learning for many-particle systems on various levels. Firstly, the node-level representation learning can be used to predict the properties of individual particles, e.g., the atomic energy, dipoles and forces. A particular application is to fit QM PES for atomistic systems, where the energy of the whole system is decomposed into contributions from individual atoms. One can employ Molecular CT to learn a proper node embedding $\mathbf{n}_i^{(T)}$ for the $i$-th atom, then predict the target property (i.e., the energy) based on this representation using a node-level readout model, e.g. a neural network or kernel method. In the experiments, we adopted a multi-layer perceptron (MLP) $f_\theta^{(\mathcal{N})}$ as the node-level readout function to predict the atomic energy $E_i$ and forces $\mathbf{F}_i$, the loss function of which is given by Eq. (14), where $E_0$ is the label for the total energy of all atoms, $\mathbf{F}_{i,0}$ is the label for the atomic force and $\lambda$ is a hyper-parameter balancing the training signals from energy and forces. By minimizing this loss, Molecular CT along with the readout function $f_\theta^{(\mathcal{N})}$ gets optimized together, and we can obtain a model which learns a good representation for the atomic environment and predicts the target property well.

$$E_{i,\theta} = f_\theta^{(\mathcal{N})}\left(\mathbf{n}_i^{(T)}\right)$$

$$\mathbf{F}_{i,\theta} = -\nabla_{\mathbf{x}_i} f_\theta^{(\mathcal{N})}\left(\mathbf{n}_i^{(T)}\right) \qquad (14)$$

$$Loss = \left(1-\lambda\right)\left(E_0 - \sum_{i=1,N} E_{i,\theta}\right)^2 + \lambda \sum_{i=1,N} \left\|\mathbf{F}_{i,0} - \mathbf{F}_{i,\theta}\right\|^2$$

Secondly, Molecular CT also enables edge-level representation learning, and one can predict relational properties between any two particles using an edge-level readout function $f_\theta^{(\mathcal{E})}$. The edge-level readout function takes in the learned node vectors of a pair of particles $\left\{\mathbf{n}_i^{(T)}, \mathbf{n}_j^{(T)}\right\}$ as well as their edge vectors $\left(\mathbf{v}_{ij}, \mathbf{e}_{ij}\right)$ to make the final prediction. Usually $f_\theta^{(\mathcal{E})}$ (Eq. (15)) is invariant w.r.t. the change-of-order between the two node vectors, belonging to the family of relational neural networks[19],

$$\text{Relation}\left(\mathbf{n}_i, \mathbf{n}_j\right) = f_\theta^{(\mathcal{E})}\left(\left\{\mathbf{n}_i^{(T)}, \mathbf{n}_j^{(T)}\right\}, \mathbf{v}_{ij}, \mathbf{e}_{ij}\right) \qquad (15)$$

Edge-level representation learning can be very appealing in applications like relational reasoning between particles. For example, it is possible to predict the contact probability between two amino acid residues within a protein using the edge-level representation. The long-range couplings such as the Coulomb interactions between particles can also be approximated by the edge-level learning.

Furthermore, one can perform graph-level learning using Molecular CT as well. This can be done by introducing a graph-level readout function or "pooling" function, $f_\theta^{(\mathcal{G})}$, to aggregate all the node vectors into a single global vector $\mathbf{g}$ in a permutation-invariant manner:

$$\mathbf{g} = f_\theta^{(\mathcal{G})}\left(\left\{\mathbf{n}_1^{(T)}, \mathbf{n}_2^{(T)}, \ldots, \mathbf{n}_N^{(T)}\right\}\right) \qquad (16)$$

based on which we can make predictions about the entire system. Graph-level learning can be particularly useful in metric learning of molecules, which are widely used in clustering or dimensionality reduction of molecular conformations as well as many kernel methods.[9]

# III. EXPERIMENTS & RESULTS

In the following, we conducted a series of ablation tests for the techniques and model architectures proposed above. SchNet was chosen as a reasonable baseline because it is also GNN-based and it involves similar edge featurization and many-body operator as in Molecular CT. Therefore, by replacing the corresponding modules in SchNet by ours one by one, we can do thorough and anatomical benchmark of Molecular CT.

For a fair comparison, we first chose reasonable architecture and hyper-parameters for SchNet via cross validation. We then kept these hyper-parameters unchanged (including stacked interaction layers $T$ and embedding dimension $D$, *etc*.) when we ran our models. We adopted the default readout model throughout the experiment as suggested by SchNet.[35]

Considering that there are only few available models supporting dual-representation learning, we first compared Molecular CT with baseline models over traditional atomistic learning tasks (Eq. (13)) without any relational constraints using a well-curated dataset MD17.

We also contributed a new dataset, Ala2MM, containing both molecular structures and inter-particle relational constraints, and defined a challenging task to predict the forces and energy of a MM-described di-peptide. We showed that this dataset frustrated traditional atomistic learning methods, but can be properly tackled by Molecular CT.

MD17 dataset was split into a training set and a validation set, each containing 1024 samples; whereas Ala2MM dataset was split into a training set and a validation set, each containing 2048 samples. For both datasets, the training objective to be minimized takes the form of Eq. (14) where $\lambda = 0.99$ as suggested by previous work.[21, 22] During ablation test, we used the same training set for each model. The Adam optimizer with a learning rate of $1 \times 10^{-4}$ and default hyper-parameters was employed. We ran four independent optimizations using different random seeds and model initialization for every experiment, and reported the averaged performance as well as the standard deviations accordingly.

**1. Positional edge featurization impacts model efficiency**

In GNN-based atomistic learning models, the relative positions, $r_{ij}$, are transformed into edge vectors which are later fed into the many-body operator. For instance, SchNet adopts a Gaussian smear of the pairwise distance, which is inherited and further developed by some other models.[21, 22]

However, as introduced earlier, compared to $r_{ij}$, we suppose $\log r_{ij}$ is more suitable for geometric learning because many physical interactions are expressed as power functions of $r_{ij}$ (or linear functions of $\log r_{ij}$). Besides, $\log r_{ij}$ preserves the singularity when $r_{ij}$ approaches zero. We thus performed ablation test to compare the difference between these two positional edge featurization approaches.

We implemented SchNet with $d$=32 (Eq. (1)) Gaussian radial basis functions (RBFs) of $r_{ij}$ for benchmark. The baseline SchNet model, denoted as SchNet1, has one interaction layer (i.e., CFC) which is repeated for $T = 3$ iterations (Table 1). A number of 256 convolution filters were selected through cross validation. The total number of parameters in SchNet1 model amounts moderately to ~177k (Table 1). To increase the expressivity of SchNet, we also implemented SchNet3, which consists of 3 different interaction layers (or CFCs) stacked one by one. Consequently, SchNet3 contains ~490k trainable parameters, almost triple as SchNet1 (Table 1).

To examine how $\log r_{ij}$ based edge featurization performs, we directly replaced the Gaussian RBFs in SchNet1 by our logarithm counterparts (Eq. (1)), and we named this model SchNet1+. The other settings in SchNet1+ and the training protocol were all kept the same with SchNet1. Noteworthy, SchNet1+ has the same amount of trainable parameters as SchNet1 (Table 1).

As shown in Fig. 6, SchNet1+ outperforms SchNet1 over various molecules in MD17 (including aspirin, ethanol, salicylic acid and uracil) in both training and validation, demonstrating that simply replacing $r_{ij}$ by $\log r_{ij}$ during edge featurization could improve the training efficiency of SchNet.

Moreover, the performance of SchNet1+ is even slightly better than SchNet3 in most cases, showing that the improved efficiency brought by our new edge featurization is comparable to tripling the model parameters of SchNet. Hence, this experiment showed that logarithm-based distance featurization can be directed plugged in the existing GNN-based atomistic learning models and could improve the model performance "for free".

**Table 1.** Architecture and statistics of models used in ablation tests

| Model | $T$[a] | Parameter Tying[b] | No. Parameters |
|---|---|---|---|
| SchNet1 | 3 | Yes | 177k |
| SchNet1+ | 3 | Yes | 177k |
| SchNet3 | 3 | No | 490k |
| EA1 | 3 | Yes | 91k |
| EA3 | 3 | No | 230k |
| NIU1 | N/A[c] | Yes | 91k |
| NIU3 | N/A | No | 230k |
| NIU6 | N/A | No | 441k |

a. Iterations or number of the many-body operators in Eq. (11).
b. Whether the parameters of each many-body operator is tied (or shared) with others.
c. $T$ is automatically and adaptively determined in NIUs.

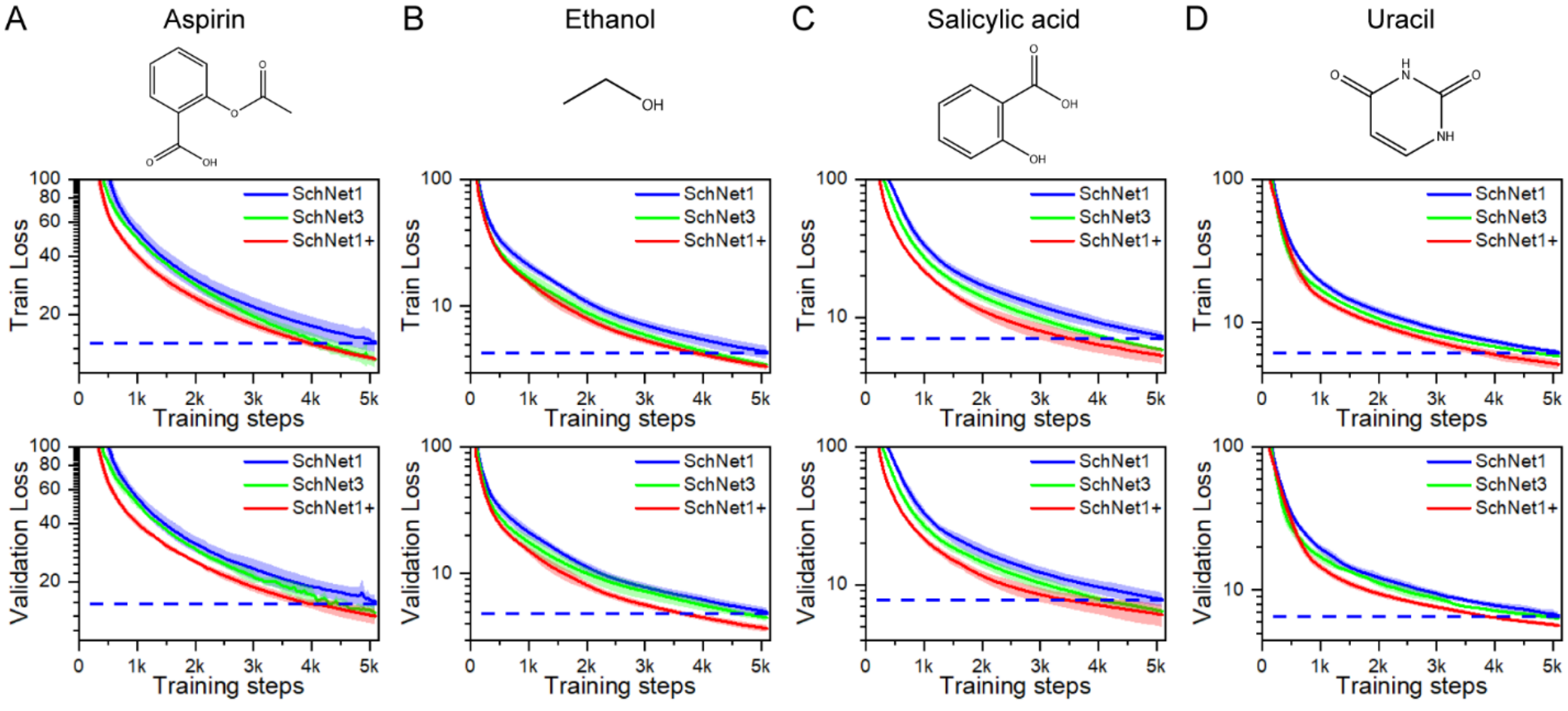

**FIGURE 6**. Ablation test for logarithm edge featurization. Loss curves for training (upper panels) and validation (lower panels) were shown for four molecules from the MD17 dataset: Aspirin (A), Ethanol (B), Salicylic acid (C) and Uracil (D), respectively. For each molecule, we performed four independent optimizations and reported the averages (solid colored lines) and standard deviations (filled shadow areas). SchNet1 (blue) and SchNet3 (green) use distance-based edge features while SchNet1+ (red) uses logarithm-distance based edge features. SchNet1 and SchNet1+ has the same number of model parameters, while SchNet3 has nearly triple number of model parameters. SchNet1 was chosen as baseline, and a blue dashed line indicating its final loss is drawn for visual guide.

**2. Ego-Attention (EA) is an expressive many-body operator**

We then benchmarked the performance of EA as a geometric many-body operator against the baseline CFC adopted by SchNet. According to the previous section, SchNet3 performed better than SchNet1, we thus chose SchNet3 as baseline for the following benchmark and ablation tests.

To compare EA with the CFC implemented in SchNet3, we tested two models. The first model (EA1) is similar to SchNet1 in that it has only one EA block which is repeated for three iterations; The other (EA3) is similar to SchNet3, equipped with three different EA blocks which are stacked sequentially. In both models, $k = 8$ heads were used for MHA (Eq. (3)).

To make a fair comparison, EA1 shares the same hyper-parameters (including the dimension of the positional edge vector $d$ and the node embedding dimension $D$) with SchNet1, whereas EA3 shares hyper-parameters with SchNet3.

However, due to the lack of the filter-generating network and the position-wise FFN in Eq. (10) (which are required by SchNet), one EA operator is roughly half the weight (i.e., number of trainable parameters) of one CFC operator. Therefore, EA1 and EA3 are only of half the size of SchNet1 and SchNet3, respectively. Both models are significantly smaller than the baseline SchNet3 (Table 1).

As shown in Fig. 7, EA1 easily exceeds the baseline SchNet3 in all the tests, showing significantly improved training efficiency during modeling of various molecules, while using only less than 20% amount of parameters than SchNet3. This experiment revealed that EA is a very efficient many-body operator, and using a single EA operator can lead to better performance than using three different CFC operators.

Furthermore, Fig. 7 also shows that by stacking more than one different EA operators, the expressivity of the overall model can be further enhanced. Noteworthy, when going “deeper” as one triples the number of many-body operators, the improvement achieved by EA3 over EA1 is much larger than that by SchNet3 over SchNet1 (Fig. 6). Therefore, EA operator can be very useful to assemble relatively “deep” model architecture.

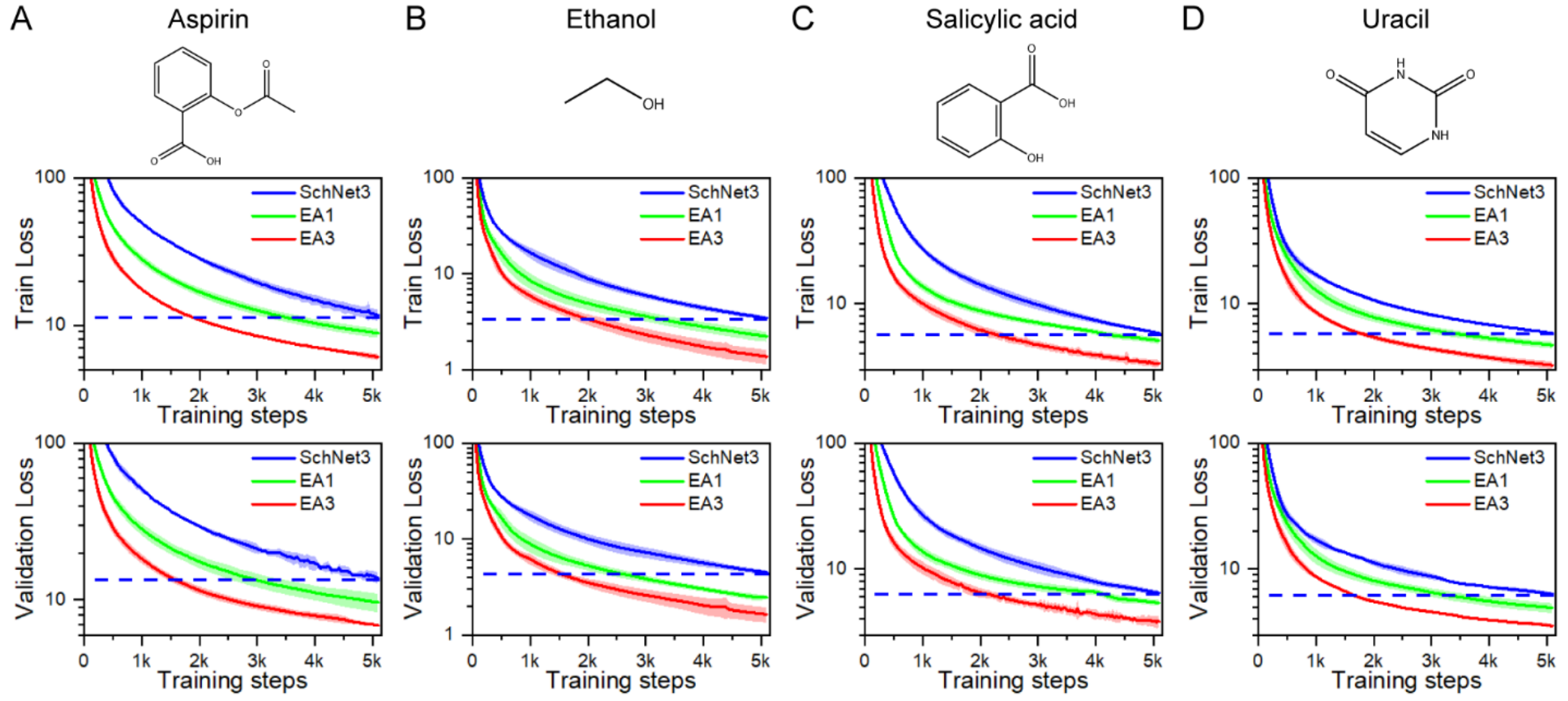

**FIGURE 7**. Ablation test for Ego-Attention. Loss curves for training (upper panels) and validation (lower panels) were shown for four molecules: Aspirin (A), Ethanol (B), Salicylic acid (C) and Uracil (D), respectively. For each molecule, we performed four independent optimizations and reported the averages (solid colored lines) and standard deviations (filled shadow areas). EA1 (green) corresponds to one EA operator repeated for three times, while EA3 (red) to three different EA operators stacked sequentially. SchNet3 (blue) is chosen as baseline for comparison, and a blue dashed line indicating its final loss is drawn for visual guide.

**3. Learning with adaptive computational complexity**

As previous section suggests, the iterations ($T$ in Eq. (11)) of a single many body operator or the number of different and stacked many-body operators (including CFC and EA) could dramatically influence the performance of GNN-based models. This issue not only leads to the computational non-universality, but also cause severe practical concerns when deploying such models

In order to address this issue, we equipped each EA operator with a pondering network, yielding a computational universal NIU. Instead of a pre-chosen fixed $T$, NIU can learn by itself how many EA operations should perform for each node.

Figure 8 shows that Molecular CT with only one NIU outperforms the baseline SchNet3 by a great margin, while using only one-fifth amount of parameters of SchNet3.

Besides, albeit Molecular CT with one NIU is theoretically Turing complete, one may still stack several different NIUs sequentially into deeper Molecular CT models. Similar to the previous experiment, we found that Molecular CT consisting of more than one NIU could lead to better training efficiency.

We also noticed that NIU consistently outperforms EA alone, although only a very limited number of extra parameters are introduced in NIU compared to EA. Particularly, when multiple NIUs are stacked together, the advantage of NIUs against stacked EAs could be more significant, provided that each NIU allows the node embedding to be refined for multiple rounds while each EA only allows once.

Last but not least, the light-weight nature of NIU allows us to build relative deep Molecular CT by stacking a number of NIUs. In the experiment, we assembled up to 6 NIUs in the largest Molecular CT (i.e., NIU6) which is at least three times deeper than SchNet3, however, its parameter number is still smaller than SchNet3 (Table 1). We found that NIU6 could achieve the best performance for all the tests presented here. In summary, the parameter-efficiency of NIUs enables us to build deeper models that have larger capacity and expressivity, thus may benefit the pre-training or representation learning over large dataset.

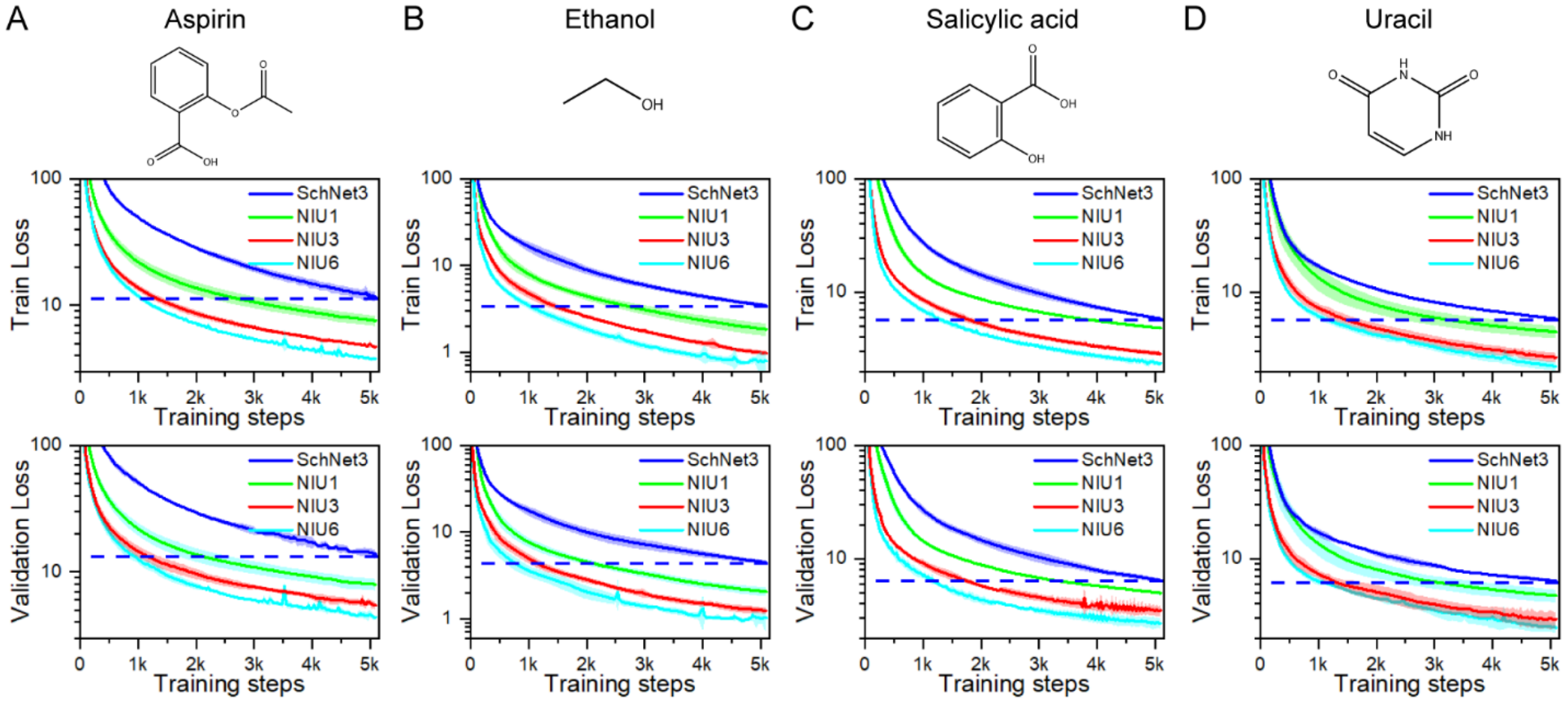

**FIGURE 8**. Ablation test for Neural Interaction Unit (NIU). Loss curves for training (upper panels) and validation (lower panels) were shown for four molecules: Aspirin (A), Ethanol (B), Salicylic acid (C) and Uracil (D), respectively. For each molecule, we performed four independent optimizations and reported the averages (solid colored lines) and standard deviations (filled shadow areas). SchNet3 (blue) is chosen as baseline for comparison, and a blue dashed line indicating its final loss is drawn for visual guide.

## 4. Learning molecular properties with relational constraints

The above experiments verified the capacity of Molecular CT in atomistic learning, i.e., learning without relational constraints between particles. However, many molecular modeling tasks impose relational constraints, and we contributed a new benchmark dataset as such.

The new dataset, Ala2MM, contains 3D structures of a terminally-blocked alanine dipeptide (Fig. 9A and 9B). For each structure, the associated label contains its potential energy and the forces of each atom. Unlike the MD17 dataset where the energy and forces were calculated by *ab initio* methods, the labels in Ala2MM dataset were obtained via MM force field. Specifically, we modeled alanine dipeptide using AMBER03 force field[36] and generated equilibrium configurations in implicit solvent[37].

MM force field differs with *ab initio* methods in that it depends on the relational constraints (namely, the bond connectivity) between particles (Fig. 9A). Therefore, to learn molecular properties determined by MM force field requires the model to be able to deal with relational constraints.

We then examined the performance of SchNet and Molecular CT on this challenging dataset. We first performed SchNet3 on Ala2MM dataset, treating atoms in alanine dipeptide quantum mechanically. In other words, the initial node embedding for each atom was directly mapped from the nuclear charge (i.e., the atomic number) corresponding to that atom.

Figure 9C shows that SchNet3 quickly led to overfitting (indicated by the obvious divergence between the validation loss and training loss), and failed on the learning task. Such failure is expected because SchNet lacks the ability to account for relational constrains and cannot distinguish atoms of the same type but with different bond connectivity.

We then manually assigned each atom in Ala2MM dataset an artificial "atomic number" according to their bonding connectivity, so that atoms can be distinguished by their associated bond types. We found that such artificially assigned atom types rescued the behavior of SchNet3 (denoted as SchNet3MM in Fig. 9C) in that no early overfitting was observed. This experiment demonstrated the relational constraints could break the built-in symmetry or invariance of atomistic learning models, and additional information must be integrated in order to capture such asymmetry for models like SchNet.[7]

On the other hand, Molecular CT is able to account for relational constraints by virtue of RME. As is seen in Fig. 9D, Molecular CT worked well on this learning tasks, and the training process is as efficient as previous atomistic learning on the QM MD17 dataset. Unlike SchNet3MM which only works when the atom types are artificially modified properly, Molecular CT only reads in the ground-true atomic number of an atom as its initial node embedding, and automatically learns to transform it into a relation-ware one via the RME, hence negating any manual interference or pre-processing. In contrast, when the RME was removed from Molecular CT, the model (denoted as MolCT/RME- in Fig. 9D) also failed the learning task, indicated by a similar early overfitting as is observed for SchNet3. This result suggests the essential role of RME in Molecular CT when dealing with many-particle systems containing relational constraints.

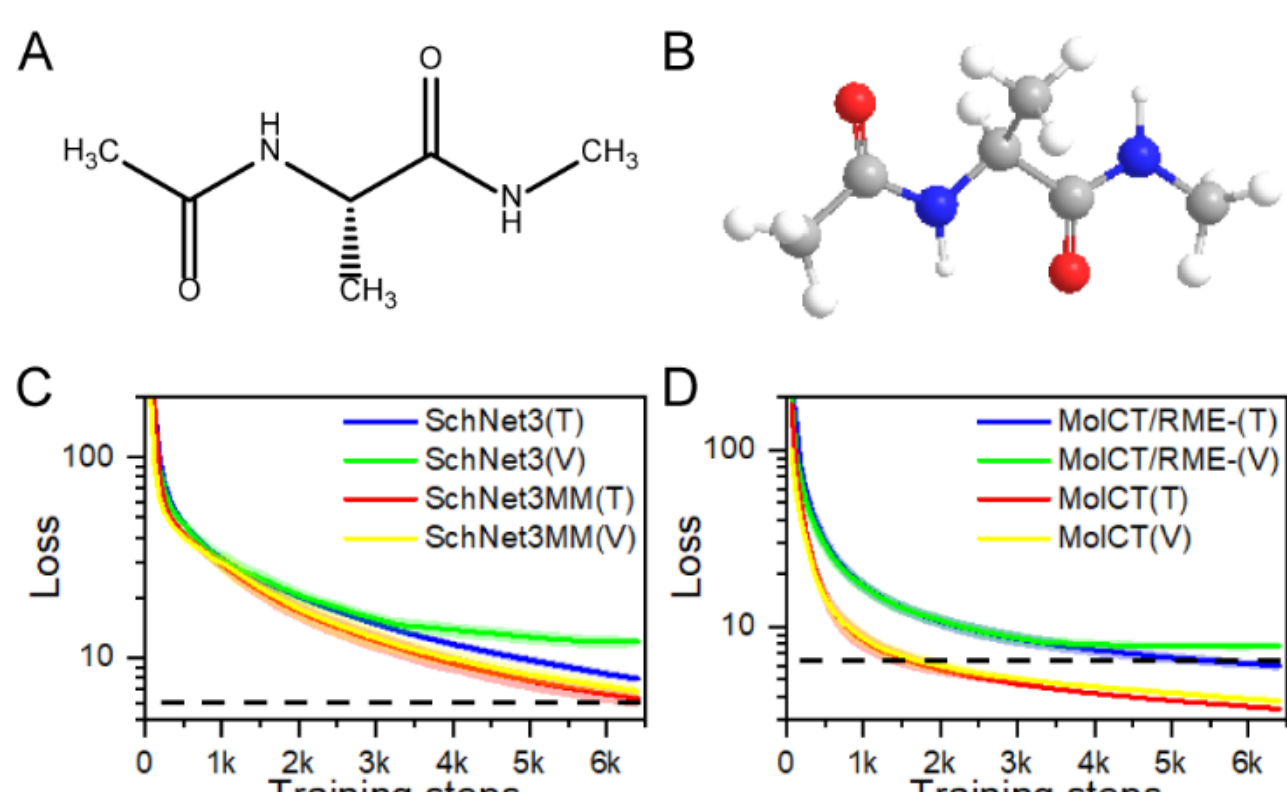

**FIGURE 9**. Benchmark Molecular CT on Ala2MM dataset. Ala2MM contains energy and forces of alanine dipeptide dependent on both the bonding connectivity (A) and the positions of atoms (B). (C) Performance of SchNet3 on Ala2MM. The suffix "(T)" and "(V)" correspond to the training and validation losses, respectively. (D) Performance of Molecular CT on Ala2MM. Molecular CT without RME are denoted as MolCT/RME-. In (C) and (D), a black dashed line indicating the final training loss of SchNet3MM is shown for visual guide.

## IV. CONCLUDING REMARKS & OUTLOOK

In this work, we introduced a new deep neural network architecture, Molecular CT, for general-purpose learning of many-particle systems. Given the dual representations of the many-particle systems, Molecular CT is invariant w.r.t. the trans-rotational transform of the many-particle system, meanwhile is able to account for any relational constraints that influence the properties of the system.

In Molecular CT, we represent the many-particle systems by a featurized graph based on physics and chemistry motivations. Particularly, the distances between particles are first transformed into the logarithm-scale, and mapped into an information-rich vector as the relative positional embedding inspired by Transformer. The relational constraints between particles can also be embedded in a similar way. As a result, Molecular CT allows one to model atoms, molecules and even coarse-grained particles like backbones of proteins within a unified architecture. This is one major advantage of Molecular CT over those models designed only for QM-level atomistic learning.

Equipped with EA, an efficient many-body operator for geometric learning, Molecular CT is very expressive in terms of approximating or learning the many-body structural patterns. Therefore, as shown in the ablation tests, compared to other models, Molecular CT is able to achieve comparable or even better performance at much lighter weight, thus providing additional ease of implementation for users to train and run the model in practice. We also bypass the stationary assumption in most of the GNN-based models by designing the NIU. Established upon NIUs, Molecular CT is computationally universal and Turing complete, and it performs adaptive representation learning for particles of different types or under different contexts. This attribute is essential for transfer or meta learning between different systems. Unlike most methods treating the neural networks as black-box approximators, by inspecting the ego-attention coefficients and the adaptive computational time as learned by Molecular CT, we may get a clue on how the particles interact with each other.

Moreover, by virtue of the GNN-based design, Molecular CT enables representation learning of many-particle systems on various levels. For example, in this paper, we show that on the node-level, Molecular CT can fit atomistic forces at both QM and MM accuracies. Edge-level and graph-level learning based on Molecular CT are left for further research.

As the EA mechanism is computationally efficient and suitable for automatic parallelism, Molecular CT can be lifted to heavier weight if needed. For example, the Transformer has been scaled up by many variants such as BERT[38], GPT-3[39] *etc.* for pre-training and transfer learning which revolutionized natural language modeling. Inspired by these successes and given that Molecular CT is a universal representation learner, it is appealing to pre-train Molecular

`

CT on a large scale and enable transfer learning for downstream tasks where the labels are expensive.

As in other atomistic learning models, the current Molecular CT adopts a finite-horizon approximation based on a cutoff function. Although this assumption is reasonable in most cases, it could be non-trivial to choose a good cutoff boundary which trades-off the computational budget and the model performance during transferable learning. While most of the current atomistic learning models suffer issues of scalability with respect to. larger cutoffs, the attention-based Molecular CT is likely to overcome this issue provided with the recent advance of the generalized sparse attention mechanisms[40] and auto-parallel machine learning technique supported by new deep learning frameworks like MindSpore[41]. We leave this idea for future research.

The demand for proper deep neural network models compatible with atomistic and molecular systems is on a surge, as there is increasing evidence that deep learning may help address challenging molecular modeling problems, e.g., solving the Schrodinger equation, predicting the protein structures, coarse-graining biomolecules, sampling molecular structures, and extracting chemical reaction mechanisms, etc. We expect Molecular CT to serve as a useful tool for all these exciting research fields given its great versatility and scalability.

## CODE AVAILABLE

We developed a general program library *Cybertron* for GNN-based deep molecular models based on Huawei's all-scenario AI framework MindSpore (www.mindspore.cn). Currently it has three built-in models: Molecular CT, SchNet and PhysNet. *Cybertron* can be downloaded from this website: gitee.com/helloyesterday/cybertron.

## ACKNOWLEDGEMENTS

The authors thank Dr. Xing Che and Dr. Lijiang Yang for useful discussion. This research was supported by National Key R&D Program of China (No. 2022ZD0115003), National Natural Science Foundation of China [22003042 to Y. I. Y. and 92053202, 21821004 to Y.Q.G], and Guangdong Basic and Applied Basic Research Foundation [2019A1515110278 to Y.I.Y.].